# Delayformer: spatiotemporal transformation for predicting high-dimensional dynamics

*Zijian Wang#, Peng Tao#, Luonan Chen**

***Abstract*—Predicting time-series is of great importance in various scientific and engineering fields. However, in the context of limited and noisy data, accurately predicting dynamics of all variables in a high-dimensional system is a challenging task due to their nonlinearity and also complex interactions. Current methods including deep learning approaches often perform poorly for real-world systems under such circumstances. This study introduces the Delayformer framework for simultaneously predicting dynamics of all variables, by developing a novel multivariate spatiotemporal information (mvSTI) transformation that makes each observed variable into a delay-embedded state (vector) and further cross-learns those states from different variables. From dynamical systems viewpoint, Delayformer predicts system states rather than individual variables, thus theoretically and computationally overcoming such nonlinearity and cross-interaction problems. Specifically, it first utilizes a single shared Visual Transformer (ViT) encoder to cross-represent dynamical states from observed variables in a delay embedded form and then employs distinct linear decoders for predicting next states, i.e. equivalently predicting all original variables parallelly. By leveraging the theoretical foundations of delay embedding theory and the representational capabilities of Transformers, Delayformer outperforms current state-of-the-art methods in forecasting tasks on both synthetic and real-world datasets. Furthermore, the potential of Delayformer as a foundational time-series model is demonstrated through cross-domain forecasting tasks, highlighting its broad applicability across various scenarios.**

## I. INTRODUCTION

Time-series data, which comprises sequences of values or observations obtained over successive periods, is ubiquitous across diverse domains, such as meteorology (temperature[1], precipitation[2], typhoon movement[3]), finance (stock prices[4], economic indicators[5]), healthcare (heart rate[6], patient monitoring[7]), epidemic outbreak prediction[8], and industry (instrument detection[9], electricity allocation[10]). Modeling time-series data for the real-world systems remains a significant challenge[11, 12] despite its widespread applications, with numerous mathematical and computational approaches developed over the years. Among these, time-series forecasting (TSF) stands out as particularly critical.

In early studies, statistical analysis and machine learning approaches such as autoregressive (AR) models[13], seasonal autoregressive integrated moving average with exogenous variables (SARIMAX)[14, 15], and support vector machines (SVM)[16] were commonly employed. More recently, deep learning has shown remarkable performance in natural language processing (NLP)[17], computer vision (CV)[18], and time-series forecasting[19], attributed to its strong representation capability. In the realm of TSF, methods based on recurrent neural networks (RNNs)[20], temporal convolutional networks (TCNs)[21], linear models[22], and Transformers[23] have emerged as mainstream approaches. Notably, despite the challenges posed by linear models[22, 24], the emergence of models like PatchTST[25], Crossformer[26], and iTransformer[27] indicates the high development potential of Transformer-based models. These models highlight the importance of incorporating patching and channel-independence in Transformers for TSF[28, 29]. However, they still face challenges in learning from limited and noisy training samples, especially in high-dimensional and short-term forecasting scenarios, because they tend to overlook the dynamic properties of the underlying system. Additionally, Transformers often rely on heuristics methods to handle interactions between different variables and are less effective in embedding temporal information[30].

Dynamical system theory, which treats time-series data as observations of a system state, different from statistical theory, offers a wide range of analytical tools, including phase space reconstruction[31], Lyapunov exponents[32], attractors[33],

# Equal contribution

Z. Wang and P. Tao are with Key Laboratory of Systems Health Science of Zhejiang Province, School of Life Science, Hangzhou Institute for Advanced Study, University of Chinese Academy of Sciences, Chinese Academy of Sciences, Hangzhou 310024, China

L. Chen is with Key Laboratory of Systems Health Science of Zhejiang Province, School of Life Science, Hangzhou Institute for Advanced Study, University of Chinese Academy of Sciences, Chinese Academy of Sciences, Hangzhou 310024, China; Key Laboratory of Systems Biology, Shanghai Institute of Biochemistry and Cell Biology, Center for Excellence in Molecular Cell Science, Chinese Academy of Sciences, Shanghai 200031, China; Guangdong Institute of Intelligence Science and Technology, Hengqin, Zhuhai, Guangdong 519031, China

* Corresponding author, Luonan Chen, email: lnchen@sibcb.ac.cn

Z. Wang and P. Tao conceived the idea; Z. Wang and P. Tao designed the research; Z. Wang performed the research; All authors analyzed the data and wrote the paper.

The source code used in this work is available at https://github.com/Wangzj000/Delayformer.

bifurcation theory[34], differential equation models[35], and time-delay embedding techniques[36]. Delay embedding theory suggests that the information in the phase space (or feature space in machine learning) of a system is often redundant[37]. The fractal dimension of attractors in high-dimensional nonlinear systems tends not to be large and is actually rather low even with many variables[38]. Moreover, according to the delay embedding theory, the topology of a dynamical system can generally be reconstructed from the delay embedding of any observed variable. This allows one to transform the spatial information (the interactions between variables) of high-dimensional data into the temporal dynamics of a single variable, a process known as spatiotemporal information (STI) transformation, thus immediately leading to the prediction of this variable. Many methods have been developed based on STI for this purpose. Ma et al.[39] introduced a randomly distributed embedding (RDE) framework for one-step-ahead prediction by constructing multiple STI maps separately. Subsequently, multilayer neural networks[40], reservoir computing[41], Transformer[42], multitask Gaussian process regression[43] and temporal convolutional networks[41, 44] have been applied to enhance the prediction accuracy and/or computational efficiency. Recently, Wu et al.[45] proposed a feature-and-reconstructed manifold mapping approach that combines feature embedding and delay embedding for forecasting all components in complex systems. However, the mappings between the feature embedding and multiple delay embeddings are solved independently, which ignores their interactions or consistent information. Moreover, STI-based methods are typically tested on short time-series data, and their prediction performance on long time-series data remains unclear. In other words, accurately and simultaneously forecasting multiple variables for both short and long time-series remains a challenge due to their complex interactions and nonlinearity (Fig. 1a).

In order to accurately make multi-step predictions for all variables simultaneously, we propose Delayformer method, which actually predicts the states rather than individual variables from the observed high-dimensional data, by exploiting the advantages of both Transformer-based and STI-based methods. Specifically, Delayformer first designs a multivariate STI (mvSTI) equation which transforms each observed variable into a delay-embedded state (vector) and then solves the mapping of states between the original system and the delay-embedded system across different variables with the mvSTI equation by leveraging the powerful representational capability of Transformer. From a dynamical systems viewpoint, Delayformer predicts the states rather than individual variables, thus theoretically and computationally overcoming those nonlinearity and cross-interaction problems. Finally, Delayformer employs distinct linear decoders for predicting states, i.e. equivalently predicting all original variables parallelly. Compared to existing Transformer-based methods, Delayformer has a solid theoretical foundation based on the dynamical system theory, which endows it with superior predictive performance in high-dimension short-term forecasting scenarios. In contrast to existing STI methods, it treats the delay-embedded Hankel matrix as a two-dimensional image, where the global and local structures of the Hankel matrix respectively correspond to the long-range and short-range dependencies of the image. To simultaneously consider both the global and local structural properties, we adopted the architecture of Vision Transformer (ViT[46]) to extract common dynamical representations of delay-embedded states rather than original variables in the system, followed by distinct decoders to predict next states or equivalently all individual variables concurrently. Due to fully exploiting the advantages of both Transformers and STI methods, Delayformer not only surpasses state-of-the-art (SOTA) methods in both long-term and short-term tasks but also demonstrates robustness to noise and hyperparameters. Finally, we tested a cross-domain forecasting task, which demonstrated that Delayformer is able to exceed the best result cross-domain predictions and zero-shot predictions, thus confirming the potential of Delayformer as a foundational model for time-series prediction.

## II. Methodology

### A. Delay-embedding theorem and spatiotemporal information (STI) equation

For a general discrete-time $N$-dimensional dynamical system with evolution function $\phi: \mathbb{R}^N \to \mathbb{R}^N$, the dynamics can be written as

$$\mathbf{X}^{t+1} = \phi(\mathbf{X}^t), \tag{1}$$

where $\mathbf{X}^t = (x_1^t, x_2^t, \dots, x_N^t)^T$ ($T$ stands a transpose) represents the $N$-dimensional observation vector/state at time point $t$, sampled at equally time intervals from the system. The Takens' embedding theorem[36] is stated as:

If $\mathcal{M} \subseteq \mathbb{R}^N$ is an attractor with the box-counting dimension $d$, for a smooth diffeomorphism $\phi: \mathcal{M} \to \mathcal{M}$ and a smooth function $h: \mathcal{M} \to \mathbb{R}$, the mapping $\mathbf{\Phi}_{\phi,h}: \mathcal{M} \to \mathbb{R}^L$ is an embedding when $L > 2d$, that is

$$\begin{aligned}\mathbf{\Phi}_{\phi,h}(\mathbf{X}^t) &= \left(h(\mathbf{X}^t), h\circ\phi(\mathbf{X}^t), \cdots, h\circ\phi^{L-1}(\mathbf{X}^t)\right)^T \\ &= \left(h(\mathbf{X}^t), h(\mathbf{X}^{t+1}), \cdots, h(\mathbf{X}^{t+L-1})\right)^T,\end{aligned} \tag{2}$$

where symbol "$\circ$" represents the function composition operation.

Letting $\mathbf{\Phi}_{\phi,h} = \mathbf{\Phi}_k$ for short and $h(\mathbf{X}^t)$ to be any target variable $x_k^t$, i.e., $h(\mathbf{X}^t) = x_k^t$, then we have the primary STI equation:

$$\mathbf{\Phi}_k(\mathbf{X}^t) = (x_k^t, x_k^{t+1}, \cdots, x_k^{t+L-1})^T = \mathbf{Y}_k^t. \tag{3}$$

One can also derive its conjugate form as $\mathbf{\Psi}_k(\mathbf{Y}_k^t) = \mathbf{\Phi}_k^{-1}(\mathbf{X}^t) = \mathbf{X}^t$ since the embedding is a one-to-one mapping, satisfying $\mathbf{\Phi}_k \circ \mathbf{\Psi}_k = id$ and $id$ represents the identity function.

As theoretically derived from the delay-embedding theory, the STI equation can transform the spatial information of high-dimensional data ($\mathbf{X}^{Known} = \{\mathbf{X}^t\}_{t=M-L-m+2}^{M-L+1}$) to the temporal information ($\mathbf{Y}_k^{Mix} = \{\mathbf{Y}_k^t\}_{t=M-m+1}^{M}$) of any target variable. In detail, for multi-variable forecast tasks, let $\mathbf{X}^t =$ $(x_1^t, x_2^t, \dots, x_N^t)^T$ represent the $N$-dimensional time-series data or the original state at time $t$ where $t = 1, 2, \dots, M$, and $T$ stands for a transpose. Then, one can construct a corresponding delayed vector $\mathbf{Y}_k^t = (x_k^t, x_k^{t+1}, \dots, x_k^{t+L-1})^T$ for the $k$-th variable with $L > 0$ as the embedding dimension. According to the delay embedding theory, $\mathbf{Y}_k^t$ can be viewed as an embedded state although it is formed only by the $k$-th variable of the original state $\mathbf{X}^t$, and thus the original state $\mathbf{X}^t$ can generally be topologically reconstructed from the delayed embedded state $\mathbf{Y}_k^t$ if $L > 2d > 0$ where $d$ is the box-counting dimension of the system. The STI equation illustrates that a spatial vector-sequence ($\mathbf{X}^{Known} = \{\mathbf{X}^t\}_{t=M-m+1}^{M}$) and a temporal vector-sequence ( $\mathbf{Y}_k^{Mix} = \{\mathbf{Y}_k^t\}_{t=M-m+1}^{M}$ ) can be converted into one another (as shown in Fig. 7a in Appendix) and can be written as

$$\mathbf{\Phi}_k(\mathbf{X}^{Known}) = \mathbf{Y}_k^{Mix}, \text{or}$$
$$\mathbf{\Phi}_k(\mathbf{X}^{M-m+1}, \mathbf{X}^{M-m+2}, \dots, \mathbf{X}^{M}) = (\mathbf{Y}_k^{M-m+1}, \mathbf{Y}_k^{M-m+2}, \dots, \mathbf{Y}_k^{M}), \quad (4)$$

or

$$\mathbf{\Phi}_k \begin{pmatrix} x_1^{M-m+1} & x_1^{M-m+2} & \cdots & x_1^{M} \\ x_2^{M-m+1} & x_2^{M-m+2} & \cdots & x_2^{M} \\ \vdots & \vdots & \ddots & \vdots \\ x_N^{M-m+1} & x_N^{M-m+2} & \cdots & x_N^{M} \end{pmatrix} = \begin{pmatrix} x_k^{M-m+1} & x_k^{M-m+2} & \cdots & x_k^{M} \\ x_k^{M-m+2} & x_k^{M-m+3} & \cdots & x_k^{M+1} \\ \vdots & \vdots & \ddots & \vdots \\ x_k^{M-m+L} & x_k^{M-m+L+1} & \cdots & x_k^{M+L-1} \end{pmatrix}, \quad (5)$$

where $\mathbf{\Phi}_k$ is a nonlinear differentiable function and $M > m$. Once $\mathbf{\Phi}_k$ is determined, then one can predict the $(L-1)$ unknown future values $x_k^{M+1}, x_k^{M+2}, \dots, x_k^{M+L-1}$. Clearly, (4) or (5) of the original STI is to transform a sequence of the original states to the corresponding sequence of the embedded states by $\mathbf{\Phi}_k$, thus predicting $(L-1)$ steps of a single target variable $x_k^t$. The accuracy of prediction is dependent on the method used to solve the mapping function $\mathbf{\Phi}_k$ based on the observed time-series $\mathbf{X}^t$ with $t = 1, 2, \dots, M$.

### *B. Multivariate STI (mvSTI) equation*

However, in order to predict all variables, the mappings $\{\mathbf{\Psi}_k\}_{k=1}^{N}$ in STI equation are usually independently solved one by one, which is extreme inefficient for high-dimensional data. To make multi-step predictions for all variables simultaneously, we design a multivariate STI (mvSTI, Fig. 1b) equation, which converts the known temporal vector-sequence ( $\mathbf{Y}_k^{Known} = \{\mathbf{Y}_k^t\}_{t=M-L-m+2}^{M-L+1}$ ) into the future spatial vector-sequence ($\mathbf{X}^{Pred} = \{\mathbf{X}^t\}_{t=M+1}^{M+m}$), that is

$$\mathbf{\Psi}_k(\mathbf{Y}_k^{Knwon}) = \mathbf{X}^{Pred}, \text{or}$$
$$\mathbf{\Psi}_{\mathrm{k}}(\mathbf{Y}_k^{M-L-m+2}, \dots, \mathbf{Y}_k^{M-L+1}) = (\mathbf{X}^{M+1}, \dots, \mathbf{X}^{M+m}), \quad (6)$$

or

$$\mathbf{\Psi}_k \begin{pmatrix} x_k^{M-L-m+2} & x_k^{M-L-m+3} & \cdots & x_k^{M-L+1} \\ x_k^{M-L-m+3} & x_k^{M-L-m+4} & \cdots & x_k^{M-L+2} \\ \vdots & \vdots & \ddots & \vdots \\ x_k^{M-m+1} & x_k^{M-m+2} & \cdots & x_k^{M} \end{pmatrix} = \begin{pmatrix} x_1^{M+1} & x_1^{M+2} & \cdots & x_1^{M+m} \\ x_2^{M+1} & x_2^{M+2} & \cdots & x_2^{M+m} \\ \vdots & \vdots & \ddots & \vdots \\ x_N^{M+1} & x_N^{M+2} & \cdots & x_N^{M+m} \end{pmatrix}, \quad (7)$$

where $\mathbf{\Psi}_k$ is also a nonlinear differentiable function. Once $\mathbf{\Psi}_{\mathrm{k}}$ is determined, one can predict $m$ steps of all $N$ variables simultaneously. We can derive the mvSTI equation (6) or (7) based on the delay embedding theorem. Clearly, in contrast to (4) or (5) of the original STI, (6) or (7) of the mvSTI is to transform a sequence of the embedded states to the corresponding sequence of the original states by $\mathbf{\Psi}_k$, thus predicting $m$ steps of all variables $\mathbf{X}^t$ simultaneously. Note that (7) is able to predict all original variables $\mathbf{X}^t$ with $t = M + 1, , \dots, M + m$ from only one observed variable $x_k^t$ with $t = M - L - m + 2, , \dots, M$, which implies that we can make reliable/accurate prediction of $\mathbf{X}^t$ if using every $k = 1, 2, \dots, N$, i.e., exploring all available high-dimensional information. This is also a major feature of this method.

### *C. Solving mvSTI equation by Delayformer*

The predictive performance by straightforwardly using the mvSTI equation with only one variable $x_k^t$ is dependent on the information content of this specific $k$ and its mapping $\mathbf{\Psi}_k$, which ignores the information and interaction of other observed variables. Here, we propose Delayformer (Fig. 1c) to address these issues by exploiting the information contents and their interactions of all observed variables. Letting $\mathbf{E}$ be an encoder or function across all variables $\mathbf{X}^t$, and $\mathbf{Z}^t$ be the latent vector/matrix, then we have the following expression based on (6)

$$\mathbf{E}^{\circ}\mathbf{\Psi}_k(\mathbf{Y}_k^{M-L-m+2}, \mathbf{Y}_k^{M-L-m+3}, \dots, \mathbf{Y}_k^{M-L+1}) = \mathbf{E}(\mathbf{X}^{M+1}, \mathbf{X}^{M+2}, \dots, \mathbf{X}^{M+m}) = (\mathbf{Z}^{M+1}, \mathbf{Z}^{M+2}, \dots, \mathbf{Z}^{M+m}) \quad (8)$$

where ° denotes the function composition operation. Clearly, if we use such a single encoder $\mathbf{E}$ to cross-learn information of all observed variables with $k = 1, \dots, N$, we can obtain the reliable representation $\mathbf{Z}^t$ of $\mathbf{X}^t$ due to the consideration of all interactions and nonlinearity. Thus, with the decoder $\mathbf{D} = (\mathbf{E})^{-1}$ from $\mathbf{Z}^t$ of (5), we have the predictions $\widehat{\mathbf{X}}^t$ as

$$(\mathbf{E})^{-1}(\mathbf{Z}^{M+1}, \mathbf{Z}^{M+2}, \dots, \mathbf{Z}^{M+m}) = (\widehat{\mathbf{X}}^{M+1}, \widehat{\mathbf{X}}^{M+2}, \dots, \widehat{\mathbf{X}}^{M+m}) \quad (9)$$

In this work, we use a linear decoder $\mathbf{f}_k$ for predicting each variable $x_k$ from (5), i.e. $\widehat{\mathbf{Y}}_k^M$, thus $\mathbf{\Psi}_k = \mathbf{f}_k{}^{\circ}\mathbf{E}$ or the prediction can also be expressed as follows

$$\mathbf{f}_k(\mathbf{Z}^{M+1}, \mathbf{Z}^{M+2}, \dots, \mathbf{Z}^{M+m+1}) = \widehat{\mathbf{Y}}_k^M = (\hat{x}_k^M, \hat{x}_k^{M+1}, \dots, \hat{x}_k^{M+L})^T.$$

For efficiently training an encoder, patching has been demonstrated to be powerful for Transformer-based methods. As shown in Fig.7b in Appendix, to better embed the temporal information in Delayformer, the delay-embedded Hankel matrix of the *k*-th variable $\mathbf{Y}_k^{Knwon}$ is treated as an image. Then the matrix/image is then divided into a number of patches, which are fed into a ViT encoder $\mathbf{E}$, thus considering both global and local interactions during the training process. Note that the encoder $\mathbf{E}$ is shared among all variables, thus enabling it to cross-learn the intrinsic common representation of the dynamical system. Furthermore, since different variables correspond to different mappings $\{\mathbf{\Psi}_k\}_{k=1}^N$, the encoded common representations $\mathbf{Z}^t$ for different variables are decoded to $\left\{\mathbf{X}_k^{Pred}\right\}_{k=1}^N$ by distinct linear layers $\{\mathbf{f}_k\}_{k=1}^N$. The mapping $\mathbf{\Psi}_k = \mathbf{f}_k \circ \mathbf{E}$ can be learned through supervised training and generalized to unseen time-series.

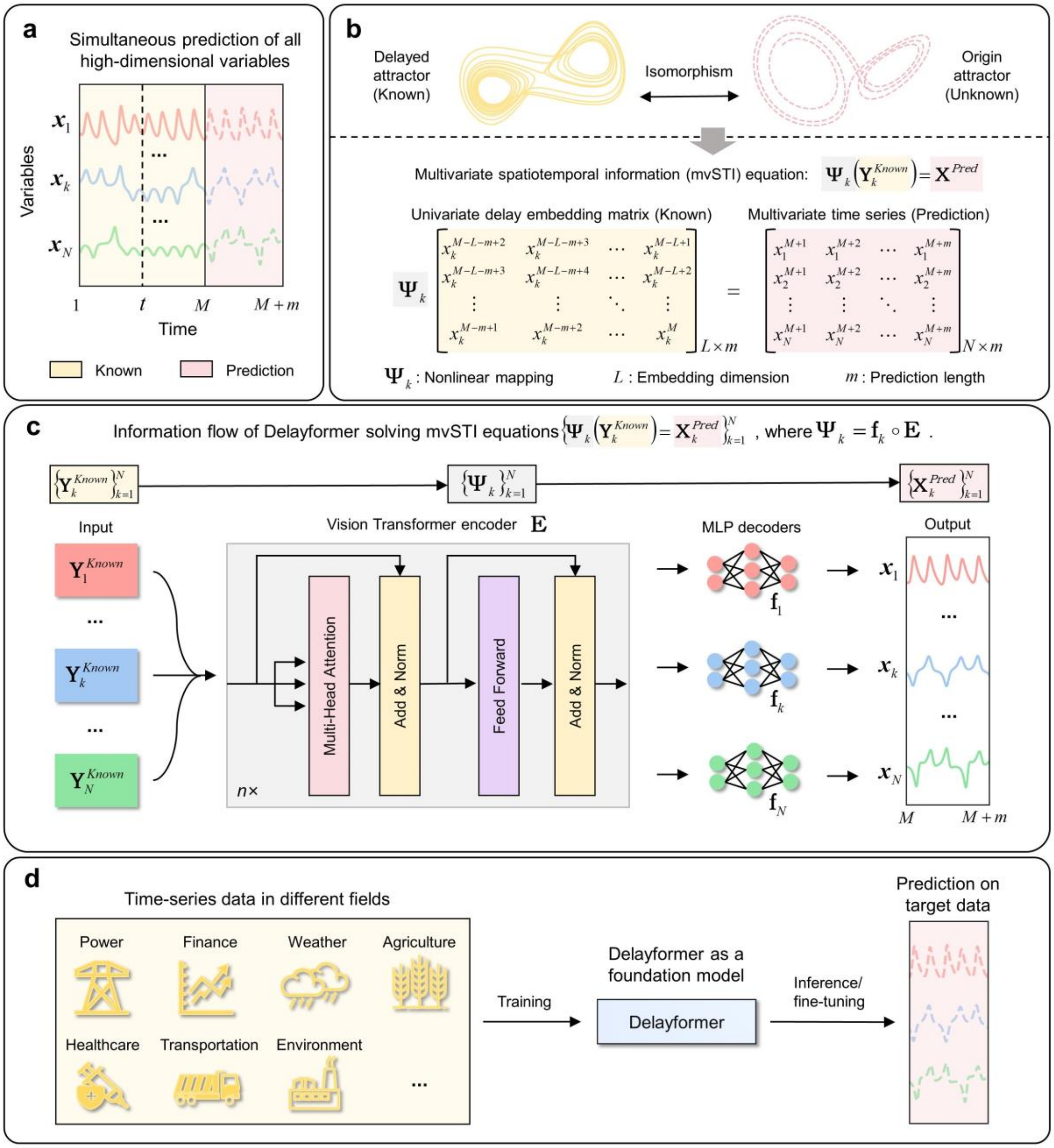

**Fig. 1. Overview of Delayformer.** **(a)** Delayformer simultaneously predicts states or all variables $(t = M + 1, , \dots, M + m)$ in a high-dimensional system based only on the observed data $(t = 1, , \dots, m)$. **(b)** In this work, we design the mvSTI equation $\mathbf{\Psi}_k(\mathbf{Y}_k^{Knwon}) = \mathbf{X}^{Pred}$ based on the delayed embedding theorem, which converts the embedded states or known temporal $(\mathbf{Y}_k^{Known} = \{\mathbf{Y}_k^t\}_{t=M-L-m+2}^{M-L+1})$ information of any variable $x_k$ into original states or the unknown future spatial information $(\mathbf{X}^{Pred} = \{\mathbf{X}^t\}_{t=M+1}^{M+m})$ of all variables, thus predicting the states or all variables simultaneously. **(c)** To solve mvSTI equation, Delayformer treats the delay-embedded Hankel matrix of the *k*-th variable $\mathbf{Y}_k^{Knwon}$ as an image, which is then divided into a number of patches and fed into a shared ViT encoder $\mathbf{E}$ for all $k = 1, \dots, N$, thus achieving the cross-learning of high-dimensional variables with the consideration of both global and local interactions. The encoded common representations $\mathbf{Z}^t$ for different variables are decoded to $\left\{\mathbf{X}_k^{Pred}\right\}_{k=1}^N$ by distinct linear layers $\{\mathbf{f}_k\}_{k=1}^N$. **(d)** Delayformer can be trained as a time-series foundation model.

## D. Implementation Details of Delayformer

*Channel-Independence*

Delayformer adopts a channel-independence (CI) strategy during training, which treats each channel (or variable) of the time-series data independently while using identical parameters for the model across channels in deep learning backbones. CI was first introduced in TSF models, demonstrating its effectiveness. Practically, CI calculates the total loss of all channels while training. Although CI exhibits low robustness on unseen variables, Delayformer enhances this approach by employing time-delay embeddings rather than the raw time-series data, thereby overcoming the robustness limitations typically associated with CI. Furthermore, the embedding vectors of each variate $\mathbf{Y}_{\boldsymbol{k}}^{\boldsymbol{Knwon}}$ for $\boldsymbol{k}=\mathbf{1},\mathbf{2},\ldots,\boldsymbol{n}$ can reconstruct the system independently, which are topologically conjugated. Then the function $\mathbf{E}$ is thus uniformly applied to represent $\mathbf{Y}_{\boldsymbol{k}}^{\boldsymbol{Knwon}}$ as $\mathbf{Z}=\mathbf{E}\left(\mathbf{Y}_{\boldsymbol{k}}^{\boldsymbol{Knwon}}\right)$ for different $\boldsymbol{k}$ across different channels, consistent with CI.

*Patching for Hankel matrixes*

Dynamical system theory illustrates that a state of the system $\mathbf{X}^t$ is primarily influenced by its proximate states. As it shown in Fig. 5a, Delayformer splits Hankel matrices into pieces of $p_1\times p_2$ as patches inspired by Vision Transformer (ViT). Like images, Hankel matrices are locally close in properties, and thus, these patches contain the local information of Hankel matrices. Since the elements of Hankel matrices are symmetric along the diagonal, $p_1$ and $p_2$ are set to be different value to confirm Delayformer can capture the features of the reconstructed systems in different fine grain sizes. Practically, the time-series for each variable with window length $W=L+m-1$, $\{x_k^t\}_{t=1}^W$, is embedded into a Hankel matrix in the shape of $L\times(W-L+1)$. This matrix is then sliced into $L\times\frac{W-L+1}{p_1p_2}$ patches, which are subsequently flattened into tokens for downstream processing. For reasons of operability, the patch number and patch dimension should be balanced to save memory while computing. In most experiments in this paper, the input length is 96, and we made embedding dimension ($L$) 49 or 27 in most cases. When $L=49$, the patch shape $(p_1,p_2)$ was often $(6,7)$ and $(24,7)$; when $L=27$, $(p_1,p_2)$ was often set to be (5,3). The detail setting for $L$ and $(p_1,p_2)$ is shown in Table III in Appendix.

*Constructing Transformer encoder*

Denoting the patches of Hankel matrices $\{\mathbf{Y}_k^{Known}\}_{k=1}^N$ as $\{P_k\}_{k=1}^N$, where each component $P_k$ contains $p=L\times\frac{W-m+1}{p_1p_2}$ patches. With the analogy to ViT, Delayformer leverages a common Transformer encoder (Vanilla Transformer) to represent the patches. The patches are projected to tokens on Transformer dimension $D$ through a linear layer after flattening, and then added a sinusoidal position encoding, i.e., the first hidden state of $k$-th variable:

$$H_k^0=\mathrm{MLP}\big(flatten(P_k)\big)+PE.$$

There are $n$ Transformer encoder blocks to calculate the correlations and representations of variate tokens. Each Transformer encoder block covers two steps, one multi-head self-attention (MSA) layer and one feed-forward (FF) layer, i.e.,

$$H_k^{j-1}=\mathrm{LayerNorm}\left(H_k^{j-1}+\mathrm{MSA}\left(H_k^{j-1}\right)\right),$$
$$H_k^{j}=\mathrm{LayerNorm}\left(H_k^{j-1}+\mathrm{FF}\left(H_k^{j-1}\right)\right).$$

The encoding representation of $k$-th variable $\mathbf{Z}_\mathrm{k}$ is the output of the last transformer encoder block $H_k^n$.

*Predicting future states and training Delayformer*

In typical TSF tasks, datasets are commonly divided into segments using sliding windows, a method that aligns well with the framework presented herein. For a dataset with batch size $B$, the input data size of one batch is $B\times W\times N$, where the input sliding window has a length $W$ and encompasses $N$ variables. Each of these $N$ variables are embedded into $L$ dimensions individually, resulting in a transformation of the batch to $B\times(W-L+1)\times L\times N$ after applying Hankelization. The resulting Hankel matrices are then treated as images, subsequently patched and flattened into tokens, yielding a shape of $B\times\frac{L\times(W-L+1)}{p_1\times p_2}\times p_1p_2\times N$. Following the structural framework established in the previous section, the tokens are first projected into $D$ dimensions, and positional encoding is added, transforming the hidden state shape of the tokens to $B\times\frac{L\times(W-L+1)}{p_1\times p_2}\times D\times N$. Each variable's tokens are then processed through $n$ Transformer encoder blocks independently, utilizing shared parameters across variables. Ultimately, $N$ bilinear layer decoders, $\{\mathbf{f}_k\}_{k=1}^N$, are employed for all variables to project the Transformer encoder representation $\{\mathbf{Z}_k\}_{k=1}^N$ into the prediction $\widehat{\mathbf{X}}^{\mathrm{pred}}$ within the feature space $\mathbb{R}^N$ for a prediction length $L$.

In this study, we employed the MSE loss to quantify the discrepancy between predicted and true values. The loss computation follows CI strategy, where the predictions of each variable are aggregated to formulate the objective function:

$$Loss=\frac{1}{M}\sum\nolimits_k\left\|\widehat{\mathbf{X}_k}^{\mathrm{pred}}-\mathbf{X}_k^{\mathrm{pred}}\right\|_2^2.$$

Adam optimizer was utilized for the model optimization. More training details shown in the Results section can be found in implementation details in Appendix.

# III. Results

## A. Performance on Lorenz systems

To validate the model performance and the fundamental mechanism of Delayformer, we first compared it with the latest competitive TSF models with Transformer backbones (iTransformer[27], PatchTST[25] and Crossformer[26]), TCN backbones (TimesNet[47]) and linear backbones (DLinear[22]) on 30-dimensional coupled Lorentz systems under various noise and parameter conditions

$$\frac{d}{dt}\mathbf{X}(t)=h\big(\mathbf{X}(t);\boldsymbol{\theta}(t)\big)+\boldsymbol{b}(t),$$

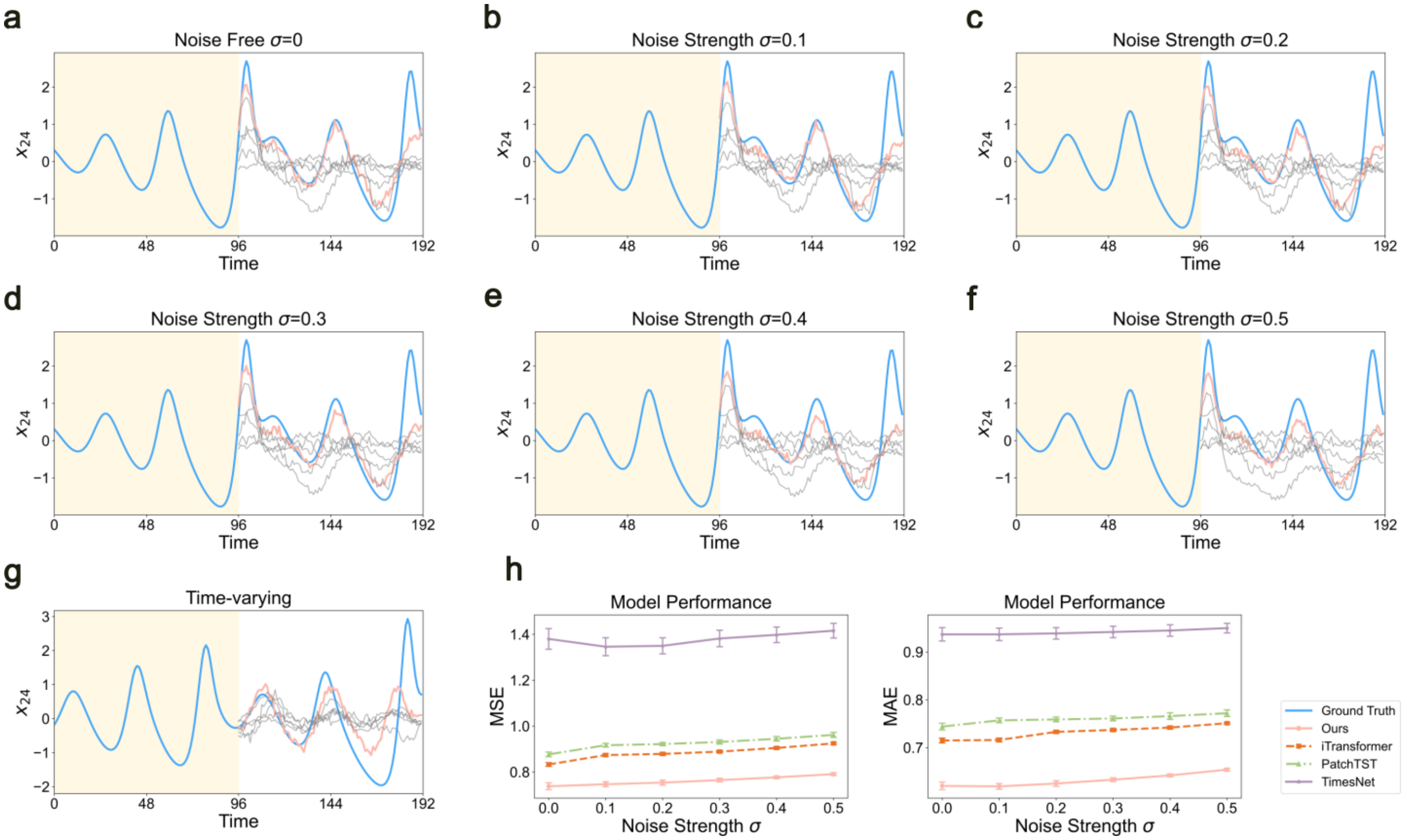

**Fig. 2. Validation of Delayformer on synthetic data.** Time-series data from a 30D coupled Lorenz system was generated under varying noise strengths and parametric conditions. Delayformer and other baselines were tested on these data. The training set, validation set, and test set sizes were 3500, 500, and 1500, respectively, for the following conditions: **(a-g)** Future state predictions (blank regions) for a sliding window of input data (shaded region) in the test set, where the blue curve represents the ground truth; the red curve represents the prediction of Delayformer; and the gray curves represent baseline models (see Figs. 8~11 in Appendix for detailed results). **(a)** Noise-free condition. **(b-f)** Conditions with additive Gaussian noise, mean value of 0, and noise strength $\sigma = 0.1, 0.2, 0.3, 0.4, 0.5$. **(g)** Time-varying condition, where the Lorenz system's parameters change over time. **(h)** Comparison of MSE and MAE for Delayformer and other baselines on the test set under different noise strengths on six identical experiments with different random number seeds, and the error bar represents the standard deviation of each condition.

where $\mathbf{X}(t) = (x_1^t, x_2^t, \dots, x_{30}^t)^T$ and $h(\cdot)$ is the nonlinear function set of the Lorentz system. $\boldsymbol{\theta}(t)$ and $\boldsymbol{b}(t)$ represent the parameter and noise vector, respectively.

*The coupled Lorenz system*

The 30-dimensional Lorentz system contains 10 3-dimensional subsystems, and the $n$-th ($n = 1,2, \dots 10$) subsystem is described as:

$$\begin{cases} \dfrac{dx_n}{dt} = \sigma(y_n - x_n) + \gamma z_{n-1}, \\ \dfrac{dy_n}{dt} = x_n(\rho - z_n) - x_n y_n, \\ \dfrac{dz_n}{dt} = -\beta z_n + x_n y_n. \end{cases}$$

$\sigma, \rho, \beta$ are the parameters of the Lorentz system, and we set them as $\sigma = 10, \rho = 28, \beta = 8/3$ under time-invariant system. $z_{n-1}$ in the third equation indicates the influence from the $(n-1)$-th system on the $n$-th system. $\gamma$ controls the interaction level between each subsystem and was set to be 0.1 constantly in the two datasets. In the time-varying Lorentz system, $\sigma$ changes as the time course: $\sigma(t) = 10 + 0.2 \times \frac{t}{10}$. The initial values are $x_n(0) = -0.1 + 0.003 \times n$, $y_n(0) = -0.097 + 0.003 \times n$, $z_n(0) = -0.094 + 0.003 \times n$.

*Time invariant and noise-free situation*

In the time-invariant and noise-free scenario, the parameter $\boldsymbol{\theta}(t)$ remains constant over time and the noise vector $\boldsymbol{b}(t)$ is set to 0. We generated 5,000 time points for the experiments, with 70% used for training (3,500 points) and the remaining 30% split evenly for validation (500 points) and testing (1,000 points). Following the current standard TSF pipeline, 96 known time points are used to predict the future unknown 96 time points. To ensure consistency in the validation environment, we set all hyperparameters in the same way (see Methods). Two metrics, mean squared error (MSE) and mean absolute error (MAE), were used to evaluate overall performance and sensitivity to outliers on the test set. Fig. 2a compares the prediction results for a randomly selected variable ($x_{24}^t$, average result is shown in Fig. 2h). The prediction of Delayformer closely matches the ground truth during multiple fluctuations within a single input sliding window, outperforming other baseline methods.

*Time-invariant and additive noise situation*

When additive Gaussian noise $\boldsymbol{b}(t)\sim N(0,\sigma^2)$ was introduced, Delayformer and five other baselines were applied to the same Lorentz system under varying noise strengths $\sigma$ (increasing from 0.1 to 0.5). Fig. 2b-f demonstrate that Delayformer effectively and consistently captures the dynamical trends and robustly anticipates future states. Moreover, although the performance of all methods decrease as the noise increases, Delayformer achieved the lowest MSE and MAE among the tested methods under different noise strengths (Fig. 2h).

*Time-varying and noise-free situation*

When the parameter vector $\boldsymbol{\theta}(t)$ varies over time (i.e., a time-varying system), the system state becomes closely related to recent short-term statuses. In this scenario, Delayformer effectively learns recent short-term information and still predicts the future states with high accuracy (MSE=0.745 and MAE=0.623), outperforming other baseline methods (Fig. 2g).

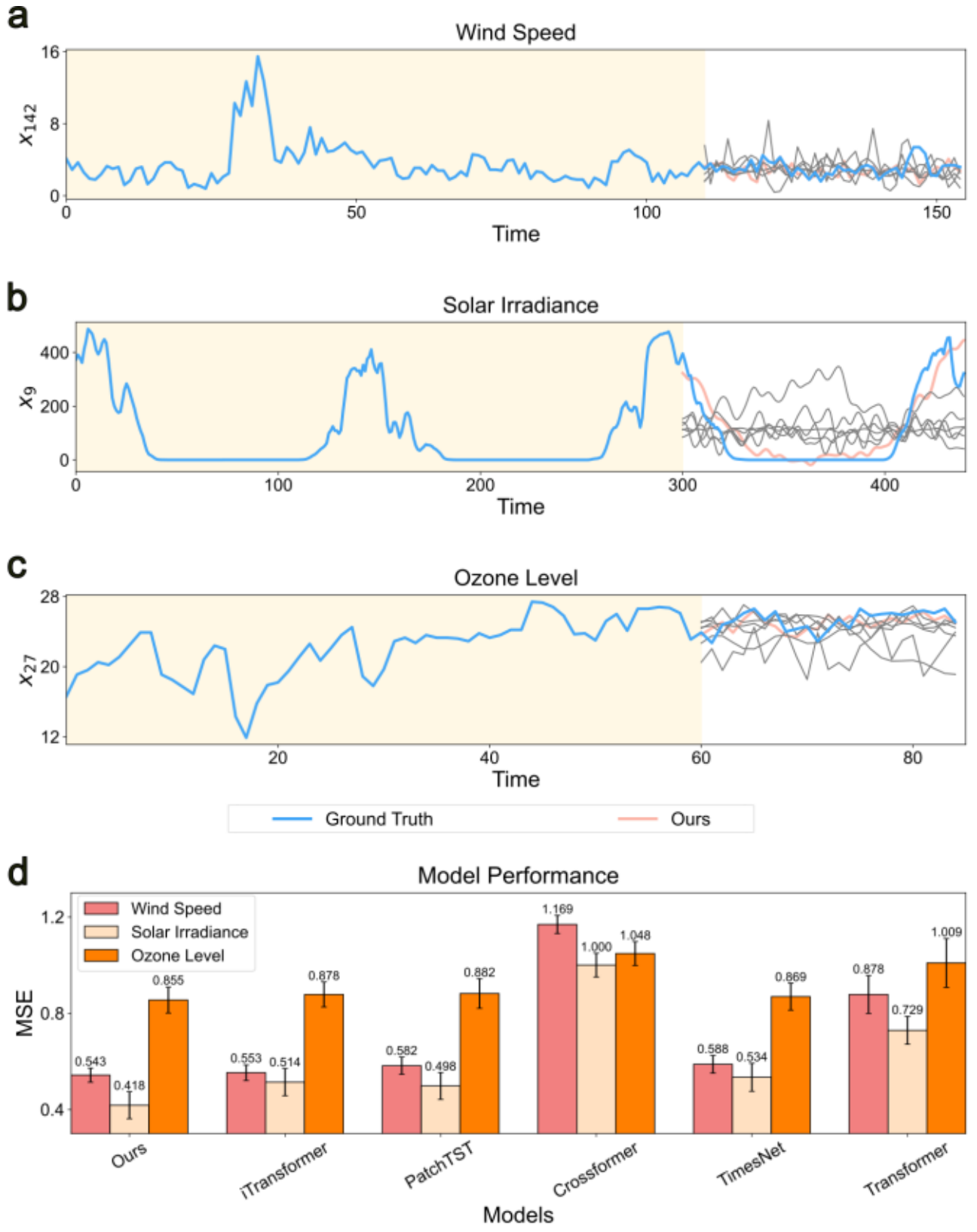

**Fig. 3. Predictions using Delayformer on three meteorological datasets.** We applied Delayformer to three meteorological datasets with limited training data. Each figure exhibits predictions of future states (blank region) based on the training data (shadow region), where the blue curve represents the ground truth, the red curve represents the prediction of Delayformer, and the gray curves represent the baselines (see Figs. 12~16 in Appendix for detailed results). **(a)** Predictions on the wind speed dataset (155 dimensions). The number of known data points is 110, and the predicted unknown data points is 45. **(b)** Predictions on the solar irradiance dataset (155 dimensions). The number of known data points is 300, and the predicted unknown data points is 140. **(c)** Predictions on the ground ozone-level dataset (72 dimensions). The number of known data points is 160, and the predicted unknown data points is 25. **(d)** Comparison of the MSE values on the three datasets for Delayformer and baseline models on six identical experiments with different random number seeds, and the error bar represents the standard error of each condition.

## B. Performance real-world short-term datasets

Previous STI-based methods, e.g., ALM and ARNN, have been tested on high-dimensional short-term datasets. Transformer-based methods usually perform poorly on these datasets due to short time-series or short-term samples. To demonstrate that Delayformer can effectively handle those datasets, we compared its performance with other Transformer-based methods (i.e., iTransformer, PatchTST, Crossformer, TimesNet, DLinear, and Transformer) on three real-world meteorological datasets. Detailed descriptions of these datasets can be found in the real-world meteorological dataset in Appendix.

*Wind speed dataset*

This 155-dimensional dataset was collected from 155 sampling sites in Wakkanani, Japan, by Japan Meteorological Agency. Using the same settings as ARNN, 110 known time points were used to train Delayformer and other baseline models to predict 45 steps ahead. As shown in Fig. 3a and d, the predictions of Delayformer are more accurate than those of other methods.

*Solar irradiance dataset*

This 155-dimensional dataset was also collected from 155 sampling sites in Wakkanani, Japan. A total of 300 known time points were used for training, with Delayformer and baseline models predicting 140 steps ahead. As illustrated in Fig. 3b and d, Delayformer outperformed the other models, yielding more precise predictions.

*Ground ozone-level dataset*

This 72-dimensional dataset was collected every hour in the Houston, Galveston, and Brazoria areas from 1998 to 2004. A total of 60 known time points were used for training, with Delayformer and baseline models predicting 25 steps ahead. Fig. 3c and d display the superior performance of Delayformer in predicting ground ozone levels.

## C. Performance on Long-term forecasting benchmarks

Previous STI-based methods have not been tested on long-term datasets. Here, to further evaluate the performance of Delayformer on general long-term forecasting, we conducted the experiments on eight benchmarks, including ETTh1[23], ETTh2[23], ETTm1[23], ETTm2[23], weather[48], solar-energy[49], electricity[48] and traffic[48]. Detailed descriptions of the datasets can be found in implementation details in Appendix.

We selected influential Transformer-based models: iTransformer[27], PatchTST[25], Crossformer[26], FEDformer[50], Stationary[51] and Autoformer[48], TCN-based models: SCINet[48] and TimesNet[47] and linear

TABLE I

RESULTS OF LONG-TERM FORECASTING BENCHMARKS. WE COMPARED OUR METHOD WITH CURRENT STATE-OF-THE-ART OR COMPETITIVE MODELS ON COMMONLY USED BENCHMARKS. "AVG" REPRESENTS THE AVERAGE METRIC OF ALL CORRESPONDING PREDICTION LENGTHS.

| Models | | Ours | | iTransformer | | Rlinear | | PatchTST | | Crossformer | | TiDE | | TimesNet | | Dlinear | | SCINet | | FEDformer | | Stationary | | Autoformer | |
|---|---|---|---|---|---|---|---|---|---|---|---|---|---|---|---|---|---|---|---|---|---|---|---|---|---|
| Metric | | MSE | MAE | MSE | MAE | MSE | MAE | MSE | MAE | MSE | MAE | MSE | MAE | MSE | MAE | MSE | MAE | MSE | MAE | MSE | MAE | MSE | MAE | MSE | MAE |
| ETTh1 | 96 | 0.379 | 0.400 | 0.386 | 0.405 | 0.386 | 0.395 | 0.414 | 0.419 | 0.423 | 0.448 | 0.479 | 0.464 | 0.384 | 0.402 | 0.386 | 0.400 | 0.654 | 0.599 | 0.376 | 0.419 | 0.513 | 0.491 | 0.449 | 0.459 |
| | 192 | 0.428 | 0.434 | 0.441 | 0.436 | 0.437 | 0.424 | 0.460 | 0.445 | 0.471 | 0.474 | 0.525 | 0.492 | 0.436 | 0.429 | 0.437 | 0.432 | 0.719 | 0.631 | 0.420 | 0.448 | 0.534 | 0.504 | 0.500 | 0.482 |
| | 336 | 0.476 | 0.451 | 0.487 | 0.458 | 0.479 | 0.446 | 0.501 | 0.466 | 0.570 | 0.546 | 0.565 | 0.515 | 0.491 | 0.469 | 0.481 | 0.459 | 0.778 | 0.659 | 0.459 | 0.465 | 0.588 | 0.535 | 0.521 | 0.496 |
| | 720 | 0.468 | 0.465 | 0.503 | 0.491 | 0.481 | 0.470 | 0.500 | 0.488 | 0.653 | 0.621 | 0.594 | 0.558 | 0.521 | 0.500 | 0.519 | 0.516 | 0.836 | 0.699 | 0.506 | 0.507 | 0.643 | 0.616 | 0.514 | 0.512 |
| | Avg | 0.438 | 0.438 | 0.454 | 0.447 | 0.446 | 0.434 | 0.469 | 0.454 | 0.529 | 0.522 | 0.541 | 0.507 | 0.458 | 0.450 | 0.456 | 0.452 | 0.747 | 0.647 | 0.440 | 0.460 | 0.570 | 0.537 | 0.496 | 0.487 |
| ETTh2 | 96 | 0.290 | 0.338 | 0.297 | 0.349 | 0.288 | 0.338 | 0.302 | 0.348 | 0.745 | 0.584 | 0.400 | 0.440 | 0.340 | 0.374 | 0.333 | 0.387 | 0.707 | 0.621 | 0.358 | 0.397 | 0.476 | 0.458 | 0.346 | 0.388 |
| | 192 | 0.369 | 0.391 | 0.380 | 0.400 | 0.374 | 0.390 | 0.388 | 0.400 | 0.877 | 0.656 | 0.528 | 0.509 | 0.402 | 0.414 | 0.477 | 0.476 | 0.860 | 0.689 | 0.429 | 0.439 | 0.512 | 0.493 | 0.456 | 0.452 |
| | 336 | 0.423 | 0.427 | 0.428 | 0.432 | 0.415 | 0.426 | 0.426 | 0.433 | 1.043 | 0.731 | 0.643 | 0.571 | 0.452 | 0.452 | 0.594 | 0.541 | 1.000 | 0.744 | 0.496 | 0.487 | 0.552 | 0.551 | 0.482 | 0.486 |
| | 720 | 0.427 | 0.441 | 0.427 | 0.445 | 0.420 | 0.440 | 0.431 | 0.446 | 1.104 | 0.763 | 0.874 | 0.679 | 0.462 | 0.468 | 0.831 | 0.657 | 1.249 | 0.838 | 0.463 | 0.474 | 0.562 | 0.560 | 0.515 | 0.511 |
| | Avg | 0.377 | 0.399 | 0.383 | 0.407 | 0.374 | 0.398 | 0.387 | 0.407 | 0.942 | 0.684 | 0.611 | 0.550 | 0.414 | 0.427 | 0.559 | 0.515 | 0.954 | 0.723 | 0.437 | 0.449 | 0.526 | 0.516 | 0.450 | 0.459 |
| ETTm1 | 96 | 0.320 | 0.362 | 0.334 | 0.368 | 0.355 | 0.376 | 0.329 | 0.367 | 0.404 | 0.426 | 0.364 | 0.387 | 0.338 | 0.375 | 0.345 | 0.372 | 0.418 | 0.438 | 0.379 | 0.419 | 0.386 | 0.398 | 0.505 | 0.475 |
| | 192 | 0.356 | 0.382 | 0.377 | 0.391 | 0.391 | 0.392 | 0.367 | 0.385 | 0.450 | 0.451 | 0.398 | 0.404 | 0.374 | 0.387 | 0.380 | 0.389 | 0.439 | 0.450 | 0.426 | 0.441 | 0.459 | 0.444 | 0.553 | 0.496 |
| | 336 | 0.388 | 0.406 | 0.426 | 0.420 | 0.424 | 0.415 | 0.399 | 0.410 | 0.532 | 0.515 | 0.428 | 0.425 | 0.410 | 0.411 | 0.413 | 0.413 | 0.490 | 0.485 | 0.445 | 0.459 | 0.495 | 0.464 | 0.621 | 0.537 |
| | 720 | 0.445 | 0.440 | 0.491 | 0.459 | 0.487 | 0.450 | 0.454 | 0.439 | 0.666 | 0.589 | 0.487 | 0.461 | 0.478 | 0.450 | 0.474 | 0.453 | 0.595 | 0.550 | 0.543 | 0.490 | 0.585 | 0.516 | 0.671 | 0.561 |
| | Avg | 0.377 | 0.398 | 0.407 | 0.410 | 0.414 | 0.407 | 0.387 | 0.400 | 0.513 | 0.496 | 0.419 | 0.419 | 0.400 | 0.406 | 0.403 | 0.407 | 0.485 | 0.481 | 0.448 | 0.452 | 0.481 | 0.456 | 0.588 | 0.517 |
| ETTm2 | 96 | 0.175 | 0.260 | 0.180 | 0.264 | 0.182 | 0.265 | 0.175 | 0.259 | 0.287 | 0.366 | 0.207 | 0.305 | 0.187 | 0.267 | 0.193 | 0.292 | 0.286 | 0.377 | 0.203 | 0.287 | 0.192 | 0.274 | 0.255 | 0.339 |
| | 192 | 0.245 | 0.308 | 0.250 | 0.309 | 0.246 | 0.304 | 0.241 | 0.302 | 0.414 | 0.492 | 0.290 | 0.364 | 0.249 | 0.309 | 0.284 | 0.362 | 0.399 | 0.445 | 0.269 | 0.328 | 0.280 | 0.339 | 0.281 | 0.340 |
| | 336 | 0.304 | 0.344 | 0.311 | 0.348 | 0.307 | 0.342 | 0.305 | 0.343 | 0.597 | 0.542 | 0.377 | 0.422 | 0.321 | 0.351 | 0.369 | 0.427 | 0.637 | 0.591 | 0.325 | 0.366 | 0.334 | 0.361 | 0.339 | 0.372 |
| | 720 | 0.401 | 0.400 | 0.412 | 0.407 | 0.407 | 0.398 | 0.402 | 0.400 | 1.730 | 1.042 | 0.558 | 0.524 | 0.408 | 0.403 | 0.554 | 0.522 | 0.960 | 0.735 | 0.421 | 0.415 | 0.417 | 0.413 | 0.433 | 0.432 |
| | Avg | 0.281 | 0.328 | 0.288 | 0.332 | 0.286 | 0.327 | 0.281 | 0.326 | 0.757 | 0.610 | 0.358 | 0.404 | 0.291 | 0.333 | 0.350 | 0.401 | 0.571 | 0.537 | 0.305 | 0.349 | 0.306 | 0.347 | 0.327 | 0.371 |
| Weather | 96 | 0.160 | 0.210 | 0.174 | 0.214 | 0.192 | 0.232 | 0.177 | 0.218 | 0.158 | 0.230 | 0.202 | 0.261 | 0.172 | 0.220 | 0.196 | 0.255 | 0.221 | 0.306 | 0.217 | 0.296 | 0.173 | 0.223 | 0.266 | 0.336 |
| | 192 | 0.208 | 0.252 | 0.221 | 0.254 | 0.240 | 0.271 | 0.225 | 0.259 | 0.206 | 0.277 | 0.242 | 0.298 | 0.219 | 0.261 | 0.237 | 0.296 | 0.261 | 0.340 | 0.276 | 0.336 | 0.245 | 0.285 | 0.307 | 0.367 |
| | 336 | 0.265 | 0.293 | 0.278 | 0.296 | 0.292 | 0.307 | 0.278 | 0.297 | 0.272 | 0.335 | 0.287 | 0.335 | 0.280 | 0.306 | 0.283 | 0.335 | 0.309 | 0.378 | 0.339 | 0.380 | 0.321 | 0.338 | 0.359 | 0.395 |
| | 720 | 0.346 | 0.346 | 0.358 | 0.349 | 0.364 | 0.353 | 0.354 | 0.348 | 0.398 | 0.418 | 0.351 | 0.386 | 0.365 | 0.359 | 0.345 | 0.381 | 0.377 | 0.427 | 0.403 | 0.428 | 0.414 | 0.410 | 0.419 | 0.428 |
| | Avg | 0.245 | 0.275 | 0.258 | 0.279 | 0.272 | 0.291 | 0.259 | 0.281 | 0.259 | 0.315 | 0.271 | 0.320 | 0.259 | 0.287 | 0.265 | 0.317 | 0.292 | 0.363 | 0.309 | 0.360 | 0.288 | 0.314 | 0.338 | 0.382 |
| Solar-Energy | 96 | 0.213 | 0.271 | 0.203 | 0.237 | 0.322 | 0.339 | 0.234 | 0.286 | 0.310 | 0.331 | 0.312 | 0.399 | 0.250 | 0.292 | 0.290 | 0.378 | 0.237 | 0.344 | 0.242 | 0.342 | 0.215 | 0.249 | 0.884 | 0.711 |
| | 192 | 0.246 | 0.295 | 0.233 | 0.261 | 0.359 | 0.356 | 0.267 | 0.310 | 0.734 | 0.725 | 0.339 | 0.416 | 0.296 | 0.318 | 0.320 | 0.398 | 0.280 | 0.380 | 0.285 | 0.380 | 0.254 | 0.272 | 0.834 | 0.692 |
| | 336 | 0.266 | 0.304 | 0.248 | 0.273 | 0.397 | 0.369 | 0.290 | 0.315 | 0.750 | 0.735 | 0.368 | 0.430 | 0.319 | 0.330 | 0.353 | 0.415 | 0.304 | 0.389 | 0.282 | 0.376 | 0.290 | 0.296 | 0.941 | 0.723 |
| | 720 | 0.267 | 0.309 | 0.249 | 0.275 | 0.397 | 0.356 | 0.289 | 0.317 | 0.769 | 0.765 | 0.370 | 0.425 | 0.338 | 0.337 | 0.356 | 0.413 | 0.308 | 0.388 | 0.357 | 0.427 | 0.285 | 0.295 | 0.882 | 0.717 |
| | Avg | 0.248 | 0.295 | 0.233 | 0.262 | 0.369 | 0.356 | 0.270 | 0.307 | 0.641 | 0.639 | 0.347 | 0.417 | 0.301 | 0.319 | 0.330 | 0.401 | 0.282 | 0.375 | 0.291 | 0.381 | 0.261 | 0.381 | 0.885 | 0.711 |
| Electricity | 96 | 0.158 | 0.260 | 0.148 | 0.240 | 0.201 | 0.281 | 0.195 | 0.285 | 0.219 | 0.314 | 0.237 | 0.329 | 0.168 | 0.272 | 0.197 | 0.282 | 0.247 | 0.345 | 0.193 | 0.308 | 0.169 | 0.273 | 0.201 | 0.317 |
| | 192 | 0.172 | 0.271 | 0.162 | 0.253 | 0.201 | 0.283 | 0.199 | 0.289 | 0.231 | 0.322 | 0.236 | 0.330 | 0.184 | 0.289 | 0.196 | 0.285 | 0.257 | 0.355 | 0.201 | 0.315 | 0.182 | 0.286 | 0.222 | 0.334 |
| | 336 | 0.191 | 0.289 | 0.178 | 0.269 | 0.215 | 0.298 | 0.215 | 0.305 | 0.246 | 0.337 | 0.249 | 0.344 | 0.198 | 0.300 | 0.209 | 0.301 | 0.269 | 0.369 | 0.214 | 0.329 | 0.200 | 0.304 | 0.231 | 0.338 |
| | 720 | 0.230 | 0.323 | 0.225 | 0.317 | 0.257 | 0.331 | 0.256 | 0.337 | 0.280 | 0.363 | 0.284 | 0.373 | 0.220 | 0.320 | 0.245 | 0.333 | 0.299 | 0.390 | 0.246 | 0.355 | 0.222 | 0.321 | 0.254 | 0.361 |
| | Avg | 0.188 | 0.286 | 0.178 | 0.270 | 0.219 | 0.298 | 0.216 | 0.304 | 0.244 | 0.334 | 0.251 | 0.344 | 0.192 | 0.295 | 0.212 | 0.300 | 0.268 | 0.365 | 0.214 | 0.327 | 0.193 | 0.296 | 0.227 | 0.338 |
| Traffic | 96 | 0.504 | 0.316 | 0.395 | 0.268 | 0.649 | 0.389 | 0.544 | 0.359 | 0.522 | 0.290 | 0.805 | 0.493 | 0.593 | 0.321 | 0.650 | 0.396 | 0.788 | 0.499 | 0.587 | 0.366 | 0.612 | 0.338 | 0.613 | 0.388 |
| | 192 | 0.513 | 0.319 | 0.417 | 0.276 | 0.601 | 0.366 | 0.540 | 0.354 | 0.530 | 0.293 | 0.756 | 0.474 | 0.617 | 0.336 | 0.598 | 0.370 | 0.789 | 0.505 | 0.604 | 0.373 | 0.613 | 0.340 | 0.616 | 0.382 |
| | 336 | 0.510 | 0.309 | 0.433 | 0.283 | 0.609 | 0.369 | 0.551 | 0.358 | 0.558 | 0.305 | 0.762 | 0.477 | 0.629 | 0.336 | 0.605 | 0.373 | 0.797 | 0.508 | 0.621 | 0.383 | 0.618 | 0.328 | 0.622 | 0.337 |
| | 720 | 0.547 | 0.332 | 0.467 | 0.302 | 0.647 | 0.387 | 0.586 | 0.375 | 0.589 | 0.328 | 0.719 | 0.449 | 0.640 | 0.350 | 0.645 | 0.394 | 0.841 | 0.523 | 0.626 | 0.382 | 0.653 | 0.355 | 0.660 | 0.408 |
| | Avg | 0.519 | 0.319 | 0.428 | 0.282 | 0.626 | 0.378 | 0.555 | 0.362 | 0.550 | 0.304 | 0.760 | 0.473 | 0.620 | 0.336 | 0.625 | 0.383 | 0.804 | 0.509 | 0.610 | 0.376 | 0.624 | 0.340 | 0.628 | 0.379 |
| 1st+2nd | | 39 | 35 | 16 | 20 | 6 | 14 | 10 | 10 | 3 | 0 | 1 | 0 | 1 | 1 | 0 | 0 | 0 | 0 | 4 | 0 | 0 | 0 | 0 | 0 |

models: RLinear[24], DLinear[22] and TiDE[52] as our baselines. In the implement of the baselines, we referred to the Time-Series-Library (https://github.com/thuml/Time-Series-Library), setting the input sequence length for our model and baselines to 96. As shown in Table I, Delayformer achieved top-two performances in all benchmarks, demonstrating its superior capability and robustness in long-term forecasting tasks.

## *D. Performance of Delayformer with Limited Data*

*Limited training set size*

Although the effectiveness of Delayformer on both short and long-term datasets has been validated, the relationship between its performance and the size of the training set remains unknown. We began by evaluating the impact of training set size on the weather dataset. Unlike typical benchmarks, we fixed the validation and test set sizes at 1,000 and 20,000 points, respectively, and varied the training set sizes from 800 to 10,000 points. Despite the wide range in training set sizes, Delayformer consistently outperformed baseline models.

a) Small training set (800 and 2,000 points): Under extremely small training set conditions, baseline models struggled to predict the future states accurately, often failing to capture the underlying dynamics of the time series (Fig. 4d and e). In contrast, Delayformer demonstrated remarkable robustness, providing reliable forecasts even with limited data.

b) Moderate and large training set (5,000 and 10,000 points): With a more substantial training set, the performance of baseline models improved, but Delayformer maintained its superior predictive accuracy (Fig. 4f and g).

The MSE across different training set sizes further confirmed

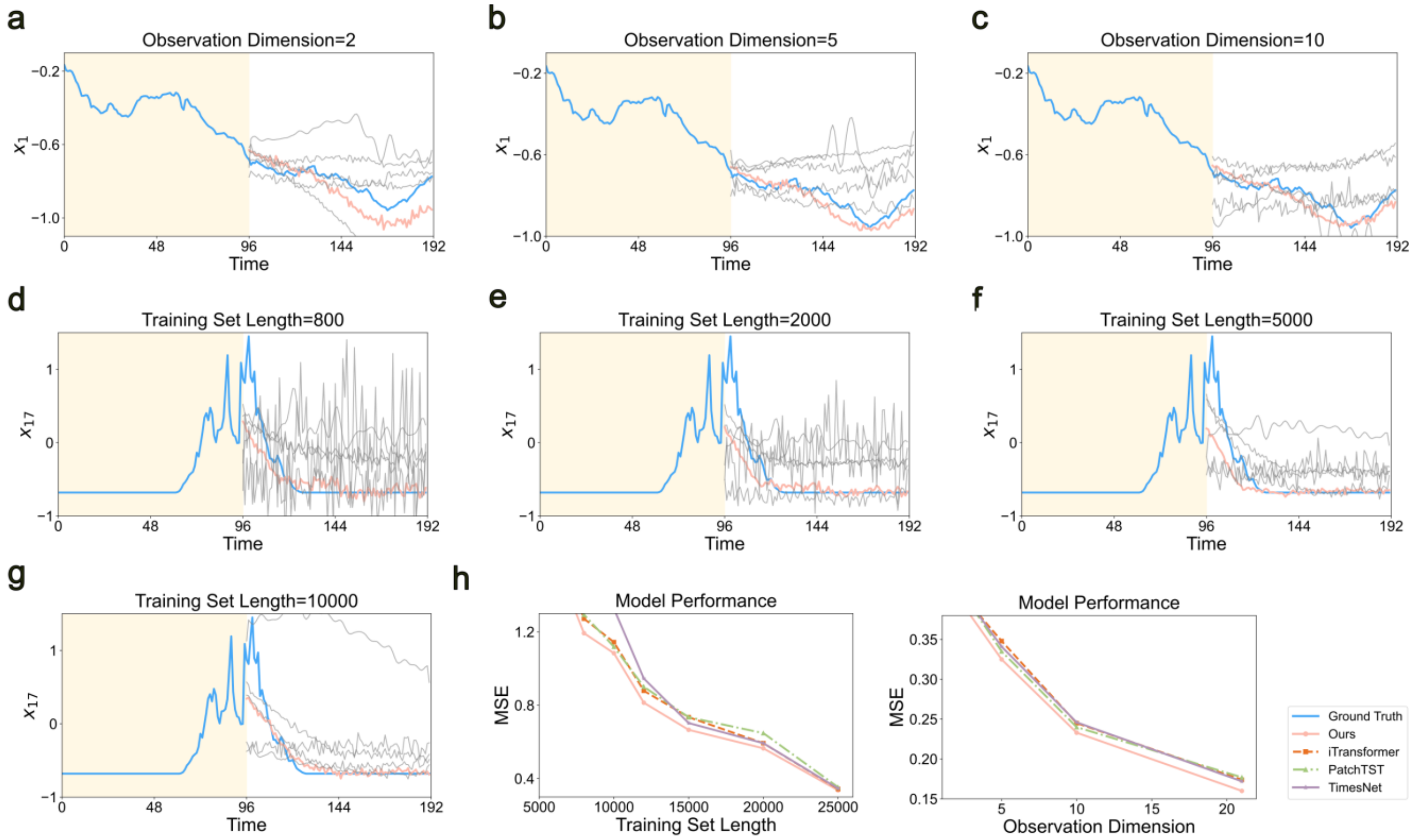

**Fig. 4. Performance of Delayformer under varying training set sizes and observation dimensions. (a-c)** Predictions of future states with different numbers of observed variables: **(a)** 2 variables, **(b)** 5 variables, **(c)** 10 variables. The shaded region represents one sliding window of input data, with the blue curve indicating the ground truth, the red curve showing Delayformer predictions, and the gray curves depicting baseline model predictions in the blank region (see Figs. 17~21 in Appendix for different baselines). **(d-g)** Predictions of future states (unshaded region) using different training set sizes: **(d)** 800 points, **(e)** 2,000 points, **(f)** 5,000 points **(g)** 10,000 points. Again, the shaded region represents one sliding window of input data, with the blue curve indicating the ground truth, the red curve showing Delayformer predictions, and the gray curves depicting baseline model predictions in the blank region. **(h)** Mean Squared Error (MSE) on the test set for Delayformer and baseline models is shown across varying training set sizes (left) and observed variables (right). Delayformer consistently exhibits lower MSE, underscoring its robustness and significant performance gains over baseline models, particularly under data-scarce conditions.

the robustness of Delayformer, consistently achieving lower MSE compared to other models (Fig. 4h).

*Limited observation dimension*

In practice, the absence of certain observed variables is common, especially in situations where only a subset of potential predictors is available. To test the ability of Delayformer to handle such scenarios, we trained the model using only the first 2, 5, and 10 variables from the dataset to predict the first variable $x_1$ for the same input sequence in the test set.

a) Few Observations (2 variables): Even with only two observed variables, Delayformer managed to extract sufficient information to predict future states effectively, outperforming the baselines (Fig. 4a).
b) Increasing Observations (5 and 10 variables): As the number of observed variables increased, the predictions of Delayformer became more accurate, further distancing itself from the baseline models (Fig. 4b and c).

The overall MSE for different observation dimensions demonstrated the capability of Delayformer to handle limited observations without significant loss of accuracy, further solidifying its robustness in practical, data-constrained environments (Fig. 4h).

## E. *Hyperparameters of Delayformer*

In Delayformer, Hankel matrices are generated by embedding one input sliding window with length $W = L + m - 1$ into $L$ dimensions, which are then sliced into patches with patching size $(p_1, p_2)$ (Fig. 5a). We conducted experiments to test the sensitivity of Delayformer to these hyperparameters $(L, p_1, p_2)$ and found that the results were relatively consistent (hyperparameter and environment used in long-term forecasting benchmark dataset in Appendix).

The embedding dimension $L$ represents the number of states in the reconstruction space. While established methods[39] provide rules for determining $L$, but in our tests, it was often calculated as an extreme value. Therefore, we set $L$ as a value between 1 and the input sequence length choosing a value that was neither too large nor too small based on empirical experience. We tested $m$ values of 12, 27, and 42 on the weather dataset (corresponding to the task in Table I) with a fixed patching size $(p_1, p_2) = (5, 3)$. The input sequence length and output sequence length were both set to 96. The predictions (Fig. 5b) and metrics (Table IV in Appendix) indicate that the results were similar across these cases.

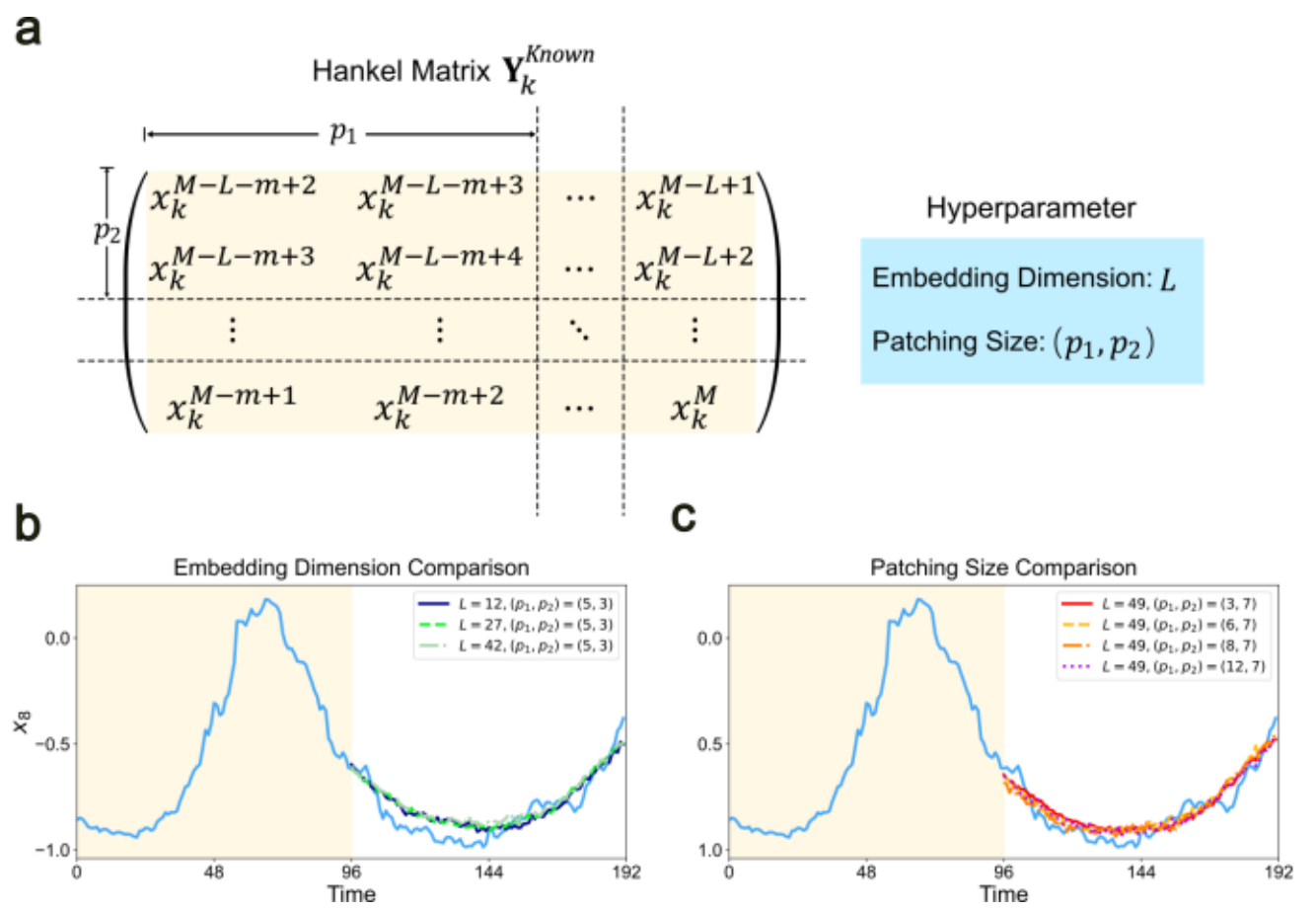

**Fig. 5. Additional hyperparameters of Delayformer. (a)** Delayformer introduces hyperparameters embedding dimension $m$ and patching size $(p_1, p_2)$. **(b)** Prediction of Delayformer trained with $L = 12, 27, 42$ with a fixed patching size of $(p_1, p_2) = (5, 3)$. The blue curve represents the ground truth, and other curves represent the predictions. **(c)** Prediction of Delayformer trained with $(p_1, p_2) = (3, 7)$, $(6, 7)$, $(8, 7)$, $(12, 7)$ with a fixed embedding dimension of $L = 49$. Again, the blue curve and other curves represent the ground truth and the predictions respectively.

The patching size $(p_1, p_2)$ determines the size of the Hankel matrix slices, representing the fine-grain size sensing range of the Hankel matrices. $(p_1, p_2)$ must satisfy the condition that $m$ is divisible by $p_1$ and $L$ is divisible by $p_2$. We tested different $(p_1, p_2)$ values on the weather dataset with $L = 49$ fixed and compared four cases: $(p_1, p_2) = (3, 7)$, $(6, 7)$, $(8, 7)$ and $(12, 7)$. The input sequence length and output sequence length were both set to 96. The predictions (Fig. 5c) and metrics (Table V in Appendix) indicate that the results were again similar across these cases.

### *F. Potential of Delayformer as a Time-Series Foundation Model*

Delayformer presents an efficient approach to constructing input tokens and cross-learning all observed variables on multi-domain tasks from time-series sequences, thereby enabling its scalability for large-scale model training (Fig. 1d). Due to space constraints, we provide two cases to illustrate the potential of Delayformer as a time-series foundation model. In contrast to prior experiments, we expanded the network depth to 10 encoder layers, aligning it with current large-scale time-series models[53, 54]. For the TSF task, we evaluated the domain generalization capabilities of Delayformer on unseen datasets, both within the same domain and across different domains from the training data. To ensure comparability with Table I, we set both the input and output sequence lengths to 96. Further details regarding hyperparameters and training are provided in Appendix.

*Case 1. In-domain generalization*

To demonstrate in-domain generalization, we selected four datasets from the energy domain: ETTh1, ETTh2, ETTm1 for training, and ETTm2 for testing. We first assessed the zero-shot performance of the pretrained Delayformer, which achieved MSE and MAE values of 0.139 and 0.254, respectively. As shown in Table I, these values are lower than the current SOTA, indicating strong in-domain generalization. Additionally, we fine-tuned the pretrained Delayformer using various proportions of available training samples from the unseen dataset (Fig. 6a). Notably, only 5% of the available training data was sufficient for the large Delayformer to achieve significantly improved performance (MSE 0.139 → 0.126, Table VI in Appendix), demonstrating its few-shot learning capabilities and ability to capture intrinsic patterns within the same domain.

*Case2. Out-of-domain generalization*

To evaluate out-of-domain generalization, we trained Delayformer on four energy domain datasets (ETTh1, ETTh2, ETTm1, ETTm2) and tested it on a weather dataset from a different domain. The zero-shot forecasting performance resulted in MSE and MAE values of 0.243 and 0.282, respectively, which are comparable to current small models as shown in Table I. These results are competitive with those of current exclusively trained small models, even though the weather dataset is outside the domain of the training data. Fine-tuning with various proportions of available training samples (Fig. 6b) revealed that only 2% of the training samples were necessary for the large Delayformer to further improve its performance (MSE 0.243 → 0.186, Table VI in Appendix). This highlights its few-shot learning capabilities and ability to transfer knowledge across domains.

Furthermore, we trained a Delayformer model on the Unified Time Series Dataset (USTD)[55], a pretraining dataset comprising 12B time points. In contrast to the configuration presented in Table I, the model consists of 10 Transformer encoder blocks, each with 12 attention heads. The input sequence length was set to 512. Due to space limitations, we present only the zero-shot prediction results on benchmark datasets in Table II.

TABLE II
ZERO-SHOT PREDICTION ON BENCHMARK DATASETS. THE DELAYFORMER WAS TRAINED ON A LARGE PRETRAINING DATASET AND EVALUATED FOR ZERO-SHOT PREDICTION PERFORMANCE.

| Zero-shot prediction (MSE) | |
|---|---|
| ETTh1 | 0.579 |
| ETTh2 | 0.219 |
| ETTm1 | 0.797 |
| ETTm2 | 0.172 |
| Weather | 0.207 |
| Electricity | 0.171 |

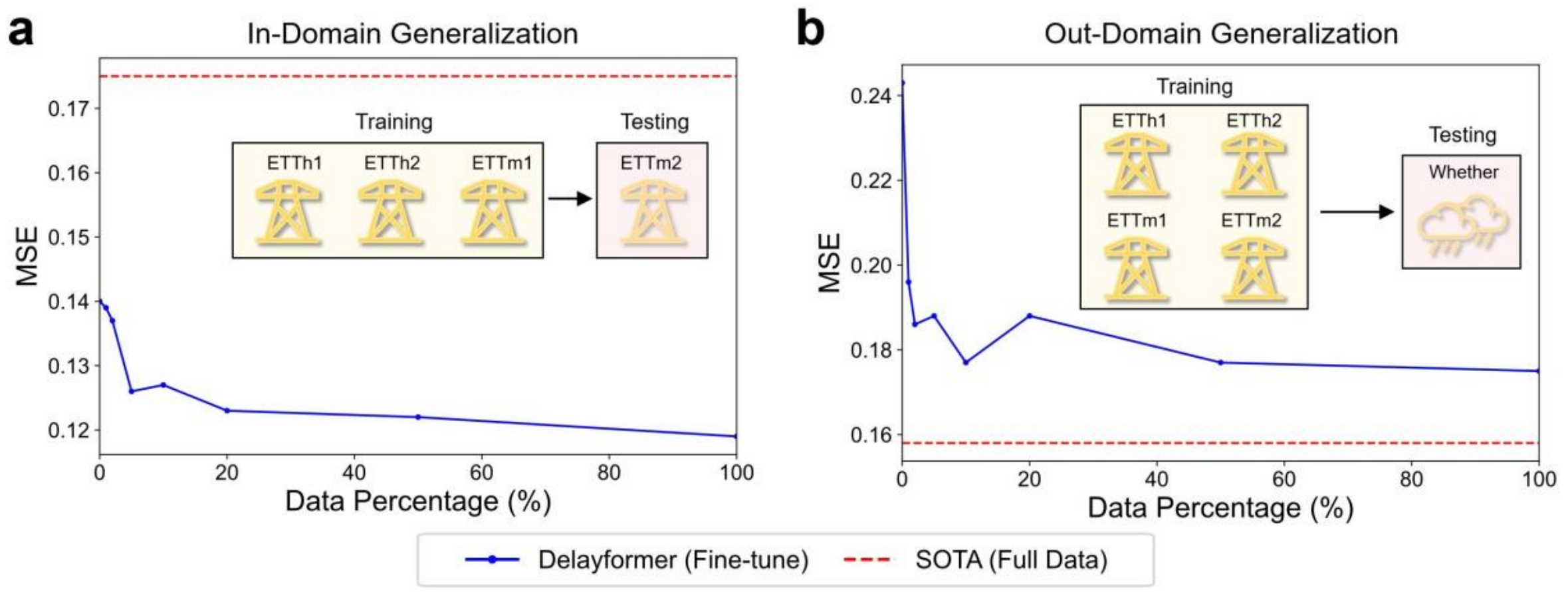

**Fig. 6. Performance of large Delayformer fine-tuned on the pretrained model.** MSE is shown for two cases, each utilizing different fine-tuning data ratios applied to the pretrained Delayformer. The blue line denotes the fine-tuned Delayformer, while the red line represents SOTA model trained with the complete set of training samples. **(a)** In-domain generalization: Training is conducted on ETTh1, ETTh2, and ETTm1 from the energy domain, with ETTm2 from the same domain as the unseen testing dataset. **(b)** Out-of-domain generalization: Training encompasses ETTh1, ETTh2, ETTm1, and ETTm2 from the energy domain, with unseen testing performed on a dataset from the weather domain, a distinct natural domain.

In conclusion, Delayformer exhibits promising scalability to match the size of current large-scale time-series models while maintaining impressive generalization capabilities. This suggests its potential for development as a foundational model for time-series.

## IV. Discussion

In this work, we proposed the Delayformer method, which is to accurately predict high-dimensional time-series by overcoming the limitations of the existing methods for handling limited and noisy data[40, 41, 45]. Specifically, this study introduces the Delayformer framework for simultaneously predicting the dynamics of all variables by developing a novel multivariate spatiotemporal information (mvSTI) transformation. This transformation converts each observed variable into a delay-embedded state/vector and cross-learns these states from different variables. From a dynamical systems viewpoint, Delayformer predicts system states rather than individual variables, thus theoretically and computationally overcoming issues related to nonlinearity and cross-interactions. Computationally, Delayformer first transforms each observed variable into a delay-embedded state (vector) by a multivariate STI (mvSTI) equation, and then solves the mapping of states between the original system and the delay-embedded system across different variables with the mvSTI equation, thus leading to the robust prediction due to the cross-learning of all variables on the same shared encoder with the consideration of both global and local interactions. By leveraging the theoretical foundations of delay embedding theory and the representational capabilities of Transformers, Delayformer outperforms current state-of-the-art methods in forecasting tasks on both synthetic and real-world datasets. Furthermore, the potential of Delayformer as a foundational time-series model is demonstrated through cross-domain forecasting tasks, highlighting its broad applicability across various scenarios.

From a computational viewpoint, Delayformer treats the Hankel matrix obtained from time-delayed embedding as a two-dimensional image and extracts the dynamical representations of all variables using a shared ViT encoder, thus overcoming the lack of temporal encoding capability of current Transformer-based methods. Notably, since the dimension of the shared encoder is relatively large, the representation ($\mathbf{Z}_k$) can be treated as the projections in the Koopman space. Koopman theory[56] suggests that the evolution of a system can be represented by a linear operator in the Koopman space. Therefore, the forecast of the system can be implemented by a linear combination of the observation functions in Koopman space.

Due to its full utilization of the strengths of both Transformers and STI methods, our experimental results demonstrate that Delayformer consistently outperforms existing models across a variety of benchmarks, including synthetic and real-world datasets. Furthermore, Delayformer exhibits robustness in scenarios with limited data and varying noise levels, making it a versatile tool for real-world applications. Finally, the introduction of hyperparameters for the embedding dimension and patching size offers additional flexibility, enabling Delayformer to be adapted to various backbones, such as decoder-only Transformers and TCNs[21], and is potential as a foundational time-series model. Thus, Delayformer provides a new approach that effectively combines deep learning with dynamical system theory, representing a significant advancement in the field of TSF.

Although Delayformer has achieved excellent prediction results on a range of benchmark datasets, it still has room for continued optimization. First, the construction of the Hankel matrix relies on two parameters, the embedding dimension and the time delay of the phase space reconstruction[57], and debugging these parameters on very large-scale datasets may impose an additional computational burden. Second, while ViT is a representative architecture used in Delayformer, it is not the SOTA architecture in the field of computer vision. More advanced model can be adopted, e.g., Convolutions Vision Transformer (CvT)[58], to further improve the prediction performance of Delayformer. Finally, although we have validated the potential of Delayformer as a large model for time-series in a cross-domain prediction task, it has not yet been trained on a sufficiently large time-series dataset, e.g. TimesFM[59] and Moment[60], which will be the focus of our future work.

In summary, by leveraging the rich theoretical framework of nonlinear dynamics and the powerful representation capability of the Transformer model, Delayformer opens a new avenue for enhancing the accuracy and robustness of time-series forecasting, thus will benefit many real-world applications.

## Acknowledgment

This work has been supported by National Key R&D Program of China (2022YFA1004800), Strategic Priority Research Program of the Chinese Academy of Sciences (XDB38040400), Natural Science Foundation of China (31930022, 12131020, T2341007, T2350003), Science and Technology Commission of Shanghai Municipality (23JS1401300), and JST Moonshot R&D(JPMJMS2021).

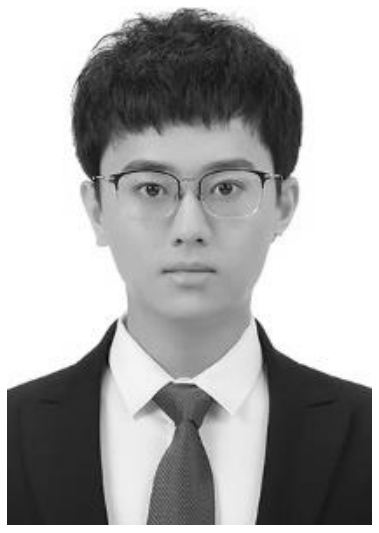

**Zijian Wang** received his B.S. degree in physics from Xiamen University in 2021. He is currently a Ph.D. candidate in Hangzhou Institute for Advanced Study, university of Chinese Academy of Science. His research interests include time series analysis, non-linear system and brain-like algorithms.

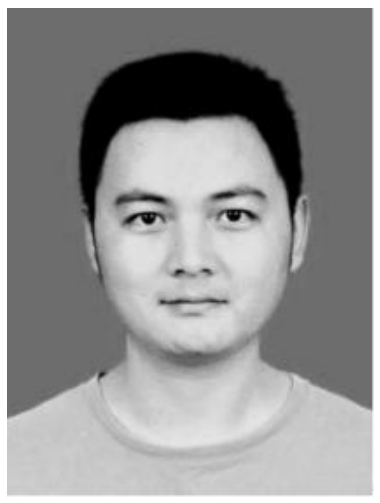

**Peng Tao** received the PhD degree from Huazhong University of Science and Technology in 2020. He is currently a research assistant at Hangzhou Institute for Advanced Study, University of Chinese Academy of Sciences, Chinese Academy of Sciences, Hangzhou, China. His research interests include in vivo folding mechanisms of biomolecules, development of machine learning methods, and design of brain-inspired learning algorithms for artificial neural networks.

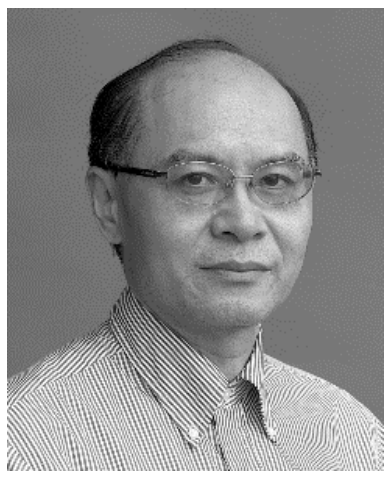

**Luonan Chen** received the BS degree from the Huazhong University of Science and Technology, Wuhan, China, in 1984, and the MS and PhD degrees from Tohoku University, Sendai, Japan, in 1988 and 1991, respectively. Since 2010, he has been a professor and executive director with the Key Laboratory of Systems Biology, Shanghai Institutes for Biological Sciences, Chinese Academy of Sciences. His fields of interests are computational systems biology, bioinformatics, and nonlinear dynamics. In recent years, he published more than 400 journal papers and two monographs in the area of computational systems biology.